\documentclass{article}
\usepackage[eabstract]{log_2024}
\usepackage{colortbl}
\usepackage{booktabs}						
\usepackage{multirow}						
\usepackage{amsfonts}						
\usepackage{graphicx}						
\usepackage{duckuments}						
\usepackage{xurl}                           
\usepackage[numbers,compress,sort]{natbib}	

\title[Recurrent Aggregators in Neural Algorithmic Reasoning]{Recurrent Aggregators in Neural Algorithmic Reasoning}

\author[Kaijia Xu and Petar Veli\v{c}kovi\'{c}]{%
Kaijia Xu\\
University of Cambridge\\
\email{kx233@cantab.ac.uk}\And
Petar Veli\v{c}kovi\'{c}\\
Google DeepMind\\
\email{petarv@google.com}
}

\begin{document}

\maketitle

\begin{abstract}

Neural algorithmic reasoning (NAR) is an emerging field that seeks to design neural networks that mimic classical algorithmic computations. Today, graph neural networks (GNNs) are widely used in neural algorithmic reasoners due to their message passing framework and permutation equivariance. In this extended abstract, we challenge this design choice, and replace the equivariant aggregation function with a \emph{recurrent neural network}. While seemingly counter-intuitive, this approach has appropriate grounding when nodes have a natural ordering---and this is the case frequently in established reasoning benchmarks like CLRS-30. Indeed, our recurrent NAR (RNAR) model performs very strongly on such tasks, while handling many others gracefully. A notable achievement of RNAR is its decisive state-of-the-art result on the Heapsort and Quickselect tasks, both deemed as a significant challenge for contemporary neural algorithmic reasoners---especially the latter, where RNAR achieves a mean micro-F$_1$ score of $87\%$.
\end{abstract}

\section{Introduction}

\textit{Neural algorithmic reasoning} \cite[NAR]{velivckovic2021neural} is an area of research that explores how \textit{neural networks} can learn \emph{algorithms} from data. This seeks to combine the benefits of both neural networks and classical algorithms and gives rise to the possibility of designing better neural networks that can learn and develop stronger algorithms for challenging real-world reasoning problems \cite{deac2021neural,velivckovic2022reasoning,he2022continuous,numeroso2023dual,georgiev2023narti,georgiev2024neural}. 

Graph neural networks \cite[GNNs]{velivckovic2023everything} are the most commonly used class of models in NAR due to  their \textit{algorithmic alignment} \cite{xu2019can} to dynamic programming \cite{dudzik2022graph}. Algorithmic alignment is the observation that an increase in the structural similarity between an algorithm and a neural network tends to result in an increase in the neural network's ability to learn the algorithm---and GNNs can offer a high degree of flexibility in how this alignment is designed \cite{xu2020neural,dudzik2024asynchronous}. Indeed, GNNs are capable of generalising out-of-distribution (OOD) on standard algorithmic benchmarks like CLRS \cite{velivckovic2022clrs} to a significantly higher degree \cite{ibarz2022generalist} than, e.g., Transformer-based LLMs \citep{markeeva2024clrs, bounsi2024transformers}.

While it is evident that this improvement is largely due to the \emph{permutation equivariance} properties of GNNs \citep{bronstein2021geometric}, so much so that often it is important for OOD generalisation to leverage strictly less expressive categories of permutation equivariant aggregators \citep{velivckovic2019neural,xu2020neural}, it is also worth noting that such an approach forces all neighbours of a node to be treated \emph{symmetrically}---and many tasks of interest to algorithms do not include such a symmetry. This is especially the case for \emph{sequential} algorithms, where the input comes in the form of a \emph{list} and hence a natural ordering between the elements exists. Indeed, such algorithms are frequent in CLRS---ten out of thirty of its tasks \cite{bentley1984programming,gavril1972algorithms,hoare1961algorithm,hoare1962quicksort,williams1964algorithm} are sequential.

In this extended abstract, we detail our attempt to leverage a \emph{recurrent} aggregator in a state-of-the-art neural algorithmic reasoning architecture (leaving all other components the same). Specifically, we leverage long short-term memory (LSTM) networks \cite{schmidhuber1997long, gers2000learning} as the aggregation function. The resulting \emph{recurrent NAR} (\textbf{RNAR}) model yielded a \emph{serendipitous} discovery: it significantly outperformed prior art on many sequential tasks in CLRS, while also gracefully handling many algorithms without such a bias! Further, RNAR sets a dominating state-of-the-art result on the Quickselect task \cite{hoare1961algorithm}, which was previously identified as a major open challenge in neural algorithmic reasoning \cite{quickselect}. 

\section{Towards RNAR}

Let $G = (V, E)$ denote a graph, where $V$ is the set of vertices and $E$ is the set of edges in the graph. Let the one-hop neighbourhoods of node $u$ be defined as $\mathcal{N}_u = \{v \in V\ |\ (v,u) \in E\}$, and $\mathbf{x}_u \in \mathbb{R}^{k}$ be the features of node $u$. With reference to the definition of GNNs in \citet{bronstein2021geometric}, we can formalise the message passing framework over this graph as: 
\begin{equation}\label{eq:message-passing-framework}
    \mathbf{x}'_{u} = \phi \left(\mathbf{x}_u, \bigoplus _{v \in \mathcal{N}_u} \psi (\mathbf{x}_u,\mathbf{x}_v)\right).
\end{equation}

The message function $\psi: \mathbb{R}^k \times \mathbb{R}^k \rightarrow \mathbb{R}^m$ first computes the message to be sent along edge $(v,u)$ based on node $u$ and its neighbour node $v$. Then, the receiver node $u$ will aggregate the messages along incoming edges using the aggregation function $\bigoplus : \mathrm{bag}(\mathbb{R}^m)\rightarrow\mathbb{R}^m$. Lastly, the update function $\phi: \mathbb{R}^k \times \mathbb{R}^m \rightarrow \mathbb{R}^k$ updates the features of the receiver node $u$ based on its current features and the aggregated messages. Typically, $\psi$ and $\phi$ are deep multilayer perceptrons.

While various forms of $\mathcal{N}_u$ have been explored by prior art \citep{velivckovic2020pointer}, nowadays it is standard practice to use a \emph{fully-connected graph} \citep{velivckovic2022clrs}, i.e. $\mathcal{N}_u=V$, and allow the GNN to infer the important neighbours by itself. This assumption also makes it easier to learn multiple algorithms in the same model \citep{ibarz2022generalist}.

\subsection{Aggregation functions}

The choice of aggregation function, $\bigoplus$, is often central to NARs' success. While it is well-known that aggregators such as summing are provably powerful for structural tasks \citep{xu2018powerful}, in practice a more aligned choice---such as $\max$---tends to be superior, especially out-of-distribution \citep{velivckovic2019neural,xu2020neural}. In nearly all cases, $\bigoplus$ is chosen to be \emph{permutation invariant}---i.e. yielding identical answers for any permutation of neighbours. Such models are known to be universal under certain conditions \citep{zaheer2017deep,wagstaff2022universal}. 

Permutation invariance is challenging to learn from data due to the high degrees-of-freedom induced by the permutation group and, as such, it is believed that this is a key reason for why GNNs tend to extrapolate better on algorithmic tasks compared to autoregressive Transformers \citep{markeeva2024clrs}. Invariance to permutations also grants the model invariance to a certain kind of asynchronous execution \citep{dudzik2024asynchronous}.

\subsection{Why would we ever \emph{drop} permutation invariance in NARs?}

With all of the above reasons in favour, it might seem extremely counter-intuitive to ever consider setting $\bigoplus$ to something which is not permutation invariant. So, \emph{why did we even bother attempting it}? 

There are three key reasons:

\begin{itemize}
    \item Firstly, permutation invariance is a property typically most desired when inputs are assumed to be given \emph{without any order}. In many algorithmic tasks, this is frequently enough not the case, making this direction worth studying. Many classical algorithm categories, such as \emph{sorting} \cite{williams1964algorithm} and \emph{searching} \cite{hoare1961algorithm}, assume that an input is a \emph{list}, inducing a natural order\footnote{We will study how important is to exploit this natural order in Appendix \ref{app:janossy}.} between the nodes. Previous research \citep{engelmayer2024parallel} highlighted how such \emph{sequential} algorithms are not favourable for GNNs.
    \item Secondly, imposing permutation symmetry forces \emph{all} neighbours to be treated \emph{equivalently}, limiting expressive power and the scope of functions that could be learnt. Recently there have been trends to eliminate various kinds of equivariances from models, leading to surprising improvements \citep{wangswallowing,abramson2024accurate}, which may also be considered motivating for our attempt.
    \item Lastly, using a permutation-invariant aggregator is typically realised by fixing a \emph{commutative monoid} structure. If the target task requires a substantially different monoid choice---often the case in more complex tasks---this can pose a unique challenge for NARs \citep{ong2022learnable}.
\end{itemize}

\subsection{The RNAR architecture}

Motivated by these reasons, in RNAR we drop the commutative monoid assumption, and instead treat $\bigoplus : \mathrm{list}(\mathbb{R}^m)\rightarrow\mathbb{R}^m$ as an arbitrary \emph{list reduction} function. We will hence assume that the $N=|V|$ node features are pre-arranged in a list $[\mathbf{x}_1, \mathbf{x}_2,\dots,\mathbf{x}_N]$. Such an ordering will always be provided by the CLRS benchmark through its \texttt{pos} node input feature \citep{velivckovic2022clrs}.

A popular, theoretically expressive choice of such a sequential aggregator is the long short-term memory (LSTM) network \citep{schmidhuber1997long}, which we employ in this work.

In a typical fully-connected GNN, each node receives messages from all other nodes, and therefore receives a total of $N$ messages in one computational step. In a GNN with an LSTM as its aggregation function, the messages from the neighbouring nodes are fed into the LSTM in a particular order. 

The LSTM will therefore run for $N$ time steps, with the input to the LSTM at each time step, $1\leq t\leq N$, being one of the $N$ messages computed using $\psi$:
\begin{equation}
    \mathbf{z}^{(u)}_t = \mathrm{LSTM}\left(\psi(\mathbf{x}_u, \mathbf{x}_t), \mathbf{z}_{t-1}^{(u)}\right)
\end{equation}
where the initial LSTM cell state, $\mathbf{z}_0^{(u)}$, is initialised to a fixed zero vector, $\mathbf{0}$. The final updated embeddings are then computed using the output of the LSTM at the last time step, which is considered to be the aggregation of the $N$ messages:
\begin{equation}
    \mathbf{x}'_u = \phi\left(\mathbf{x}_u, \mathbf{z}_N^{(u)}\right)
\end{equation}
Since the choice of the initial node ordering clearly affects $\mathbf{z}_N^{(u)}$, LSTM as an aggregator is not invariant to message receiving order, breaking permutation invariance.

While our work offers the first comprehensive study of such a recurrent aggregator on a benchmark like CLRS---and reveals surprising results---we stress that we are far from the first work to attempt replacing a GNN's aggregator with a recurrent neural network. 

Key works to consider here include GraphSAGE \citep{hamilton2017inductive}---one of the earliest GNNs to attempt an LSTM aggregator; Janossy pooling \citep{murphy2018janossy} 
and PermGNN \citep{roy2021adversarial}---illustrating how such models can be made permutation equivariant in expectation by applying them to several sampled permutations; and LCM \citep{ong2022learnable}---which showed GRU aggregators \citep{cho2014learning} can effectively learn a challenging commutative monoid. Note that RNAR uses a stronger base model than GraphSAGE; we ablate against this in Appendix \ref{app:ablation}.

\section{Evaluating RNAR}

We evaluate RNAR using the CLRS-30 algorithmic reasoning benchmark \cite{velivckovic2022clrs}. Since we want to examine whether the use of RNAR enables the emergence of novel capabilities not covered by previous state-of-the-art, we insert RNAR into the state-of-the-art Triplet-GMPNN architecture \citep{ibarz2022generalist} with hint reversals \citep{bevilacqua2023neural}, and compare its performance against the baseline Triplet-GMPNN, as well as two additional state-of-the-art neural algorithmic executors, which both offer inventive ways to boost performance: Relational Transformers \citep{diao2022relational} and G-ForgetNets \citep{bohde2024markov}. 

We remark that there are several very interesting recent works improving NARs \citep{mahdavi2022towards,rodionov2024discrete} which we exclude because they leverage a different learning regime and/or CLRS environment assumptions.

We commence with Table \ref{tab:f1-selected}, showcasing RNAR's performance on \emph{sequential} algorithms in CLRS-30, where it is expected it could perform favourably in spite of its lack of symmetry.


\begin{table}[tbh]
    \centering
    \caption{Micro-F$_1$ test OOD scores on sequential algorithms. RNAR improves on its Triplet-GMPNN baseline on 8/10 of them (underlined) and sets new state-of-the-art on 6/10.}
    \label{tab:f1-selected}
    \begin{tabular}{lcccc}
        \toprule
        \textbf{Algorithm} & \textbf{Triplet-GMPNN} & \textbf{RT} & \textbf{G-ForgetNet} & \textbf{RNAR}\\
        \midrule
        Activity Selector & $95.18\% \pm  0.45$ & $87.72\%\pm 2.7$ & $\textbf{99.03}\%\pm 0.10$ & $\underline{95.23}\% \pm  0.71$\\
        Binary Search & $77.58\% \pm  2.35$ & $81.48\%\pm 6.7$ & $\textbf{85.96}\%\pm 1.59$ & $64.71\% \pm  6.79$\\
        Bubble Sort & $80.51\% \pm 9.10$ & $38.22\% \pm 13.0$ & $83.19\% \pm 2.59$ & $\underline{\textbf{95.78}}\% \pm 0.40$\\
        Find Max. Subarray & $76.36\% \pm 0.43$ & $66.52\% \pm 3.7$ & $78.97\% \pm 0.70$ & $\underline{\textbf{83.53}}\% \pm 2.17$ \\
        Heapsort & $49.13\% \pm 10.35$ & $32.96\% \pm 14.8$ & $57.47\% \pm 6.08$ & $\underline{\textbf{93.07}}\% \pm 1.03$\\
        Insertion Sort & $87.21\% \pm 2.80$ & $89.43\% \pm 9.0$ & $\textbf{98.40}\% \pm 0.21$ & $\underline{93.00}\% \pm 1.77$\\
        Minimum & $98.43\% \pm 0.01$ & $95.28\% \pm 2.0$ & $\textbf{99.26}\% \pm 0.08$ & $96.92\% \pm 0.09$\\
        Quickselect & $0.47\% \pm 0.25$ & $19.18\% \pm 17.3$ & $6.30\% \pm 0.85$ & $\underline{\textbf{87.08}}\% \pm 2.21$\\
        Quicksort & $85.69\% \pm 4.53$ & $39.42\% \pm 13.2$ & $73.28\% \pm 6.25$ & $\underline{\textbf{94.73}}\% \pm 0.63$\\
        Task Scheduling & $87.25\% \pm  0.35$ & $82.93\%\pm 1.8$ & $84.55\%\pm 0.35$& $\underline{\textbf{88.08}}\% \pm  1.30$ \\
        \bottomrule
    \end{tabular}
\end{table}

The most important result is clearly on the \emph{Quickselect} task, wherein RNAR \textbf{sets the best recorded micro-F$_1$ score by a wide margin}, settling an important open challenge \cite{quickselect}. Besides this, RNAR is highly potent for sorting algorithms, and generally outperforms its Triplet-GMPNN baseline in nearly all of the sequential tasks.

Armed with this exciting result, we now evaluate RNAR on \emph{all} other tasks in CLRS-30 -- see Table \ref{tab:f1-all}.

\begin{table}[ht]
    \centering
    \caption{Single-task OOD average micro-F$_{1}$ score of previous SOTA: Triplet-GMPNN, RT and G-ForgetNet and our new RNAR model. For four of the algorithms (marked in red), RNAR with triplets ran out of memory on a V100 GPU, and an MPNN model \citep{gilmer2017neural} was used as a basis instead.}
    \begin{tabular}{lccccc}
    \toprule
    \textbf{Algorithm} & \textbf{Triplet-GMPNN} & \textbf{RT} & \textbf{G-ForgetNet} & \textbf{RNAR}\\\midrule
\cellcolor{green!0}Activity Selector & \cellcolor{green!0}$95.18\% \pm  0.45$ & \cellcolor{green!0}$87.72\%\pm 2.7$ & \cellcolor{green!0}$\textbf{99.03}\%\pm 0.10$ & \cellcolor{green!0}$95.23\% \pm  0.71$\\
Articulation Points & $91.04\% \pm  0.92$ & $34.15\%\pm 14.6$ & $\textbf{97.97}\%\pm 0.46$ & \cellcolor{red!30}$26.32\%\pm 27.34$\\
Bellman-Ford & $97.39\% \pm  0.19$ & $94.24\%\pm 1.5$ & $\textbf{99.18}\%\pm 0.11$ & $96.00\% \pm  0.38$\\
BFS & $ 99.93\% \pm  0.03$ & $99.14\%\pm 0.7$ & $99.96\%\pm 0.01$ & $\textbf{100.0}\%\pm 0.0$\\
\cellcolor{green!0}Binary Search & \cellcolor{green!0}$77.58\% \pm  2.35$ & \cellcolor{green!0}$81.48\%\pm 6.7$ & \cellcolor{green!0}$\textbf{85.96}\%\pm 1.59$ & \cellcolor{green!0}$64.71\% \pm  6.79$\\
Bridges & $97.70\% \pm  0.34$ & $37.88\%\pm 11.8$ & $\textbf{99.43}\%\pm 0.15$& \cellcolor{red!30}$72.22\%\pm 12.66$\\
\cellcolor{green!0}Bubble Sort & \cellcolor{green!0}$80.51\% \pm  9.10$ & \cellcolor{green!0}$38.22\%\pm 13.0$ & \cellcolor{green!0}$83.19\%\pm 2.59$ & \cellcolor{green!0}$\textbf{ 95.78}\% \pm  0.40$\\
DAG Shortest Paths & $98.19\% \pm  0.30$ & $96.61\%\pm 1.6$ & $\textbf{99.37}\%\pm 0.03$& $ 96.40\% \pm  1.47$\\
DFS & $ \textbf{100.0}\% \pm  0.00$ & $39.23\%\pm 10.5$ & $74.31\%\pm 5.03$ & $\textbf{100.0}\% \pm  0.00$\\
Dijkstra & $96.05\% \pm  0.60$ & $91.20\%\pm 5.8$ & $\textbf{99.14}\%\pm 0.06$ & $95.04\% \pm  1.62$\\
\cellcolor{green!0}Find Max. Subarray  & \cellcolor{green!0}$76.36\% \pm  0.43$ & \cellcolor{green!0}$66.52\%\pm 3.7$ & \cellcolor{green!0}$78.97\%\pm 0.70$ & \cellcolor{green!0}$\textbf{ 83.53}\% \pm  2.17$ \\
Floyd-Warshall & $48.52\% \pm  1.04$ & $31.59\%\pm 7.6$ & $\textbf{56.32}\%\pm 0.86$& $ 27.49\% \pm  6.95$\\
Graham Scan & $93.62\% \pm  0.91$ & $74.15\%\pm 7.4$ & $\textbf{97.67}\%\pm 0.14$ & $ 76.20\% \pm  4.51$\\
\cellcolor{green!0}Heapsort & \cellcolor{green!0}$49.13\% \pm 10.35$ & \cellcolor{green!0}$32.96\%\pm 14.8$ & \cellcolor{green!0}$57.47\%\pm 6.08$& \cellcolor{green!0}$\textbf{ 93.07}\% \pm  1.03$\\
\cellcolor{green!0}Insertion Sort & \cellcolor{green!0}$87.21\% \pm  2.80$ & \cellcolor{green!0}$89.43\%\pm 9.0$& \cellcolor{green!0}$\textbf{98.40}\%\pm 0.21$ & \cellcolor{green!0}$93.00\% \pm  1.77$\\
Jarvis' March & $91.01\% \pm  1.30$ & $\textbf{94.57}\%\pm 2.2$& $88.53\%\pm 2.96$ & \cellcolor{red!30}$91.83\% \pm 1.77$\\
Knuth-Morris-Pratt & $\textbf{19.51}\% \pm  4.57$ & $0.03\%\pm 0.1$ & $12.45\%\pm 3.12$ & $4.54\% \pm  2.60$\\
LCS Length & $80.51\% \pm  1.84$ & $83.32\%\pm 4.1$ & $\textbf{85.43}\%\pm 0.47$ & $66.91\% \pm  2.53$\\
Matrix Chain Order & $91.68\% \pm  0.59$ & $\textbf{91.89}\%\pm 1.2$& $91.08\%\pm 0.51$ & $ 25.12\% \pm  1.86$\\
\cellcolor{green!0}Minimum & \cellcolor{green!0}$98.43\% \pm  0.01$ & \cellcolor{green!0}$95.28\%\pm 2.0$ & \cellcolor{green!0}$\textbf{99.26}\%\pm 0.08$ & \cellcolor{green!0}$96.92\% \pm  0.09$\\
MST-Kruskal & $89.93\% \pm  0.43$ & $64.91\%\pm 11.8$ & $\textbf{91.25}\%\pm 0.40$ & \cellcolor{red!30}$67.29\% \pm 0.93$\\
MST-Prim & $87.64\% \pm  1.79$ & $85.77\%\pm 7.9$ & $\textbf{95.19}\%\pm 0.33$ & $86.60\% \pm  4.42$\\
Na\"{i}ve String Matcher & $78.67\% \pm  4.99$ & $65.01\%\pm 32.3$ & $\textbf{97.02}\%\pm 0.77$ & $93.71\% \pm  2.26$\\
Optimal BST & $73.77\% \pm  1.48$ & $74.40\%\pm 2.6$ & $\textbf{83.58}\%\pm 0.49$ & $36.04\% \pm  12.55$\\
\cellcolor{green!0}Quickselect & \cellcolor{green!0}$ 0.47\% \pm  0.25$ & \cellcolor{green!0}$19.18\%\pm 17.3$ & \cellcolor{green!0}$6.30\%\pm 0.85$ & \cellcolor{green!0}$\textbf{87.08}\% \pm  2.21$\\
\cellcolor{green!0}Quicksort & \cellcolor{green!0}$ 85.69\% \pm  4.53$ & \cellcolor{green!0}$39.42\%\pm 13.2$& \cellcolor{green!0}$73.28\%\pm 6.25$ & \cellcolor{green!0}$\textbf{94.73}\% \pm  0.63$ \\
Segments Intersect & $ 97.64\% \pm  0.09$ & $84.94\%\pm 2.6$ & $\textbf{99.06}\%\pm 0.39$  & $97.30\% \pm  0.29$\\
SCC & $43.43\% \pm  3.15$ & $28.59\%\pm 15.2$ & $\textbf{53.53}\%\pm 2.48$ & $48.43\% \pm  8.01$\\
\cellcolor{green!0}Task Scheduling & \cellcolor{green!0}$87.25\% \pm  0.35$ & \cellcolor{green!0}$82.93\%\pm 1.8$ & \cellcolor{green!0}$84.55\%\pm 0.35$& \cellcolor{green!0}$\textbf{88.08}\% \pm  1.30$ \\
Topological Sort & $87.27\% \pm  2.67$ & $80.62\%\pm 17.5$ & $\textbf{99.92}\%\pm 0.02$ & $74.00\% \pm  8.18$\\
\midrule
Overall average  & $ 80.04\%$ & $ 66.18\%$ & $ \textbf{82.89}\%$ & $75.78\%$\\
\bottomrule
\end{tabular}
\label{tab:f1-all}
\end{table}



While it is evident that removing permutation invariance does \emph{not} yield the strongest model overall, we found that performance regressions compared to Triplet-GMPNNs were not as common as expected, and only $4\%$ average performance points were lost compared to them.

Still, RNAR proves itself a worthy element in the NAR toolbox: with its outperformances on Find Max Subarray, Heapsort and especially Quickselect, there are now only \emph{three} tasks in CLRS-30 (Floyd-Warshall, Knuth-Morris-Pratt and Strongly Connected Components) for which there is no known OOD result above $80\%$---indicating that we soon may need a new test split for CLRS-30.

As such, it is our hope that RNAR inspires future research into non-commutative aggregators in NAR. We note two obvious limitations worth exploring in the future: the memory considerations of LSTM aggregators, which caused OOMs in conjunction with triplets on four of the tasks (see also Appendix \ref{app:timing}), and the fact that the Knuth-Morris-Pratt algorithm proves challenging in spite of being a string algorithm. For the former, one may consider alternatives to recurrent aggregators such as Binary-GRUs \citep{ong2022learnable}; for the latter, seeking out better alignment with \emph{automata} may be desirable.





\bibliographystyle{unsrt}
\bibliography{log_2024}

\begin{thebibliography}{46}
\providecommand{\natexlab}[1]{#1}
\providecommand{\url}[1]{\texttt{#1}}
\expandafter\ifx\csname urlstyle\endcsname\relax
  \providecommand{\doi}[1]{doi: #1}\else
  \providecommand{\doi}{doi: \begingroup \urlstyle{rm}\Url}\fi

\bibitem[Veli{\v{c}}kovi{\'c} and Blundell(2021)]{velivckovic2021neural}
Petar Veli{\v{c}}kovi{\'c} and Charles Blundell.
\newblock Neural algorithmic reasoning.
\newblock \emph{Patterns}, 2\penalty0 (7), 2021.

\bibitem[Deac et~al.(2021)Deac, Veli{\v{c}}kovi{\'c}, Milinkovic, Bacon, Tang, and Nikolic]{deac2021neural}
Andreea-Ioana Deac, Petar Veli{\v{c}}kovi{\'c}, Ognjen Milinkovic, Pierre-Luc Bacon, Jian Tang, and Mladen Nikolic.
\newblock Neural algorithmic reasoners are implicit planners.
\newblock \emph{Advances in Neural Information Processing Systems}, 34:\penalty0 15529--15542, 2021.

\bibitem[Veli{\v{c}}kovi{\'c} et~al.(2022{\natexlab{a}})Veli{\v{c}}kovi{\'c}, Bo{\v{s}}njak, Kipf, Lerchner, Hadsell, Pascanu, and Blundell]{velivckovic2022reasoning}
Petar Veli{\v{c}}kovi{\'c}, Matko Bo{\v{s}}njak, Thomas Kipf, Alexander Lerchner, Raia Hadsell, Razvan Pascanu, and Charles Blundell.
\newblock Reasoning-modulated representations.
\newblock In \emph{Learning on Graphs Conference}, pages 50--1. PMLR, 2022{\natexlab{a}}.

\bibitem[He et~al.(2022)He, Veli{\v{c}}kovi{\'c}, Li{\`o}, and Deac]{he2022continuous}
Yu~He, Petar Veli{\v{c}}kovi{\'c}, Pietro Li{\`o}, and Andreea Deac.
\newblock Continuous neural algorithmic planners.
\newblock In \emph{Learning on Graphs Conference}, pages 54--1. PMLR, 2022.

\bibitem[Numeroso et~al.(2023)Numeroso, Bacciu, and Veli{\v{c}}kovi{\'c}]{numeroso2023dual}
Danilo Numeroso, Davide Bacciu, and Petar Veli{\v{c}}kovi{\'c}.
\newblock Dual algorithmic reasoning.
\newblock \emph{arXiv preprint arXiv:2302.04496}, 2023.

\bibitem[Georgiev et~al.(2023)Georgiev, Vinas, Considine, Dumitrascu, and Lio]{georgiev2023narti}
Dobrik Georgiev, Ramon Vinas, Sam Considine, Bianca Dumitrascu, and Pietro Lio.
\newblock Narti: Neural algorithmic reasoning for trajectory inference.
\newblock In \emph{The 2023 ICML Workshop on Computational Biology}, 2023.

\bibitem[Georgiev et~al.(2024)Georgiev, Numeroso, Bacciu, and Li{\`o}]{georgiev2024neural}
Dobrik~Georgiev Georgiev, Danilo Numeroso, Davide Bacciu, and Pietro Li{\`o}.
\newblock Neural algorithmic reasoning for combinatorial optimisation.
\newblock In \emph{Learning on Graphs Conference}, pages 28--1. PMLR, 2024.

\bibitem[Veli{\v{c}}kovi{\'c}(2023)]{velivckovic2023everything}
Petar Veli{\v{c}}kovi{\'c}.
\newblock Everything is connected: Graph neural networks.
\newblock \emph{Current Opinion in Structural Biology}, 79:\penalty0 102538, 2023.

\bibitem[Xu et~al.(2019)Xu, Li, Zhang, Du, Kawarabayashi, and Jegelka]{xu2019can}
Keyulu Xu, Jingling Li, Mozhi Zhang, Simon~S Du, Ken-ichi Kawarabayashi, and Stefanie Jegelka.
\newblock What can neural networks reason about?
\newblock \emph{arXiv preprint arXiv:1905.13211}, 2019.

\bibitem[Dudzik and Veli{\v{c}}kovi{\'c}(2022)]{dudzik2022graph}
Andrew~J Dudzik and Petar Veli{\v{c}}kovi{\'c}.
\newblock Graph neural networks are dynamic programmers.
\newblock \emph{Advances in neural information processing systems}, 35:\penalty0 20635--20647, 2022.

\bibitem[Xu et~al.(2020)Xu, Zhang, Li, Du, Kawarabayashi, and Jegelka]{xu2020neural}
Keyulu Xu, Mozhi Zhang, Jingling Li, Simon~S Du, Ken-ichi Kawarabayashi, and Stefanie Jegelka.
\newblock How neural networks extrapolate: From feedforward to graph neural networks.
\newblock \emph{arXiv preprint arXiv:2009.11848}, 2020.

\bibitem[Dudzik et~al.(2024)Dudzik, von Glehn, Pascanu, and Veli{\v{c}}kovi{\'c}]{dudzik2024asynchronous}
Andrew~Joseph Dudzik, Tamara von Glehn, Razvan Pascanu, and Petar Veli{\v{c}}kovi{\'c}.
\newblock Asynchronous algorithmic alignment with cocycles.
\newblock In \emph{Learning on Graphs Conference}, pages 3--1. PMLR, 2024.

\bibitem[Veli{\v{c}}kovi{\'c} et~al.(2022{\natexlab{b}})Veli{\v{c}}kovi{\'c}, Badia, Budden, Pascanu, Banino, Dashevskiy, Hadsell, and Blundell]{velivckovic2022clrs}
Petar Veli{\v{c}}kovi{\'c}, Adri{\`a}~Puigdom{\`e}nech Badia, David Budden, Razvan Pascanu, Andrea Banino, Misha Dashevskiy, Raia Hadsell, and Charles Blundell.
\newblock The clrs algorithmic reasoning benchmark.
\newblock In \emph{International Conference on Machine Learning}, pages 22084--22102. PMLR, 2022{\natexlab{b}}.

\bibitem[Ibarz et~al.(2022)Ibarz, Kurin, Papamakarios, Nikiforou, Bennani, Csord{\'a}s, Dudzik, Bo{\v{s}}njak, Vitvitskyi, Rubanova, et~al.]{ibarz2022generalist}
Borja Ibarz, Vitaly Kurin, George Papamakarios, Kyriacos Nikiforou, Mehdi Bennani, R{\'o}bert Csord{\'a}s, Andrew~Joseph Dudzik, Matko Bo{\v{s}}njak, Alex Vitvitskyi, Yulia Rubanova, et~al.
\newblock A generalist neural algorithmic learner.
\newblock In \emph{Learning on graphs conference}, pages 2--1. PMLR, 2022.

\bibitem[Markeeva et~al.(2024)Markeeva, McLeish, Ibarz, Bounsi, Kozlova, Vitvitskyi, Blundell, Goldstein, Schwarzschild, and Veli{\v{c}}kovi{\'c}]{markeeva2024clrs}
Larisa Markeeva, Sean McLeish, Borja Ibarz, Wilfried Bounsi, Olga Kozlova, Alex Vitvitskyi, Charles Blundell, Tom Goldstein, Avi Schwarzschild, and Petar Veli{\v{c}}kovi{\'c}.
\newblock The clrs-text algorithmic reasoning language benchmark.
\newblock \emph{arXiv preprint arXiv:2406.04229}, 2024.

\bibitem[Bounsi et~al.(2024)Bounsi, Ibarz, Dudzik, Hamrick, Markeeva, Vitvitskyi, Pascanu, and Veli{\v{c}}kovi{\'c}]{bounsi2024transformers}
Wilfried Bounsi, Borja Ibarz, Andrew Dudzik, Jessica~B Hamrick, Larisa Markeeva, Alex Vitvitskyi, Razvan Pascanu, and Petar Veli{\v{c}}kovi{\'c}.
\newblock Transformers meet neural algorithmic reasoners.
\newblock \emph{arXiv preprint arXiv:2406.09308}, 2024.

\bibitem[Bronstein et~al.(2021)Bronstein, Bruna, Cohen, and Veli{\v{c}}kovi{\'c}]{bronstein2021geometric}
Michael~M Bronstein, Joan Bruna, Taco Cohen, and Petar Veli{\v{c}}kovi{\'c}.
\newblock Geometric deep learning: Grids, groups, graphs, geodesics, and gauges.
\newblock \emph{arXiv preprint arXiv:2104.13478}, 2021.

\bibitem[Veli{\v{c}}kovi{\'c} et~al.(2019)Veli{\v{c}}kovi{\'c}, Ying, Padovano, Hadsell, and Blundell]{velivckovic2019neural}
Petar Veli{\v{c}}kovi{\'c}, Rex Ying, Matilde Padovano, Raia Hadsell, and Charles Blundell.
\newblock Neural execution of graph algorithms.
\newblock \emph{arXiv preprint arXiv:1910.10593}, 2019.

\bibitem[Bentley(1984)]{bentley1984programming}
Jon Bentley.
\newblock Programming pearls: algorithm design techniques.
\newblock \emph{Communications of the ACM}, 27\penalty0 (9):\penalty0 865--873, 1984.

\bibitem[Gavril(1972)]{gavril1972algorithms}
F{\u{a}}nic{\u{a}} Gavril.
\newblock Algorithms for minimum coloring, maximum clique, minimum covering by cliques, and maximum independent set of a chordal graph.
\newblock \emph{SIAM Journal on Computing}, 1\penalty0 (2):\penalty0 180--187, 1972.

\bibitem[Hoare(1961)]{hoare1961algorithm}
Charles~AR Hoare.
\newblock Algorithm 65: find.
\newblock \emph{Communications of the ACM}, 4\penalty0 (7):\penalty0 321--322, 1961.

\bibitem[Hoare(1962)]{hoare1962quicksort}
Charles~AR Hoare.
\newblock Quicksort.
\newblock \emph{The Computer Journal}, 5\penalty0 (1):\penalty0 10--16, 1962.

\bibitem[Williams(1964)]{williams1964algorithm}
John William~Joseph Williams.
\newblock Algorithm 232: heapsort.
\newblock \emph{Commun. ACM}, 7:\penalty0 347--348, 1964.

\bibitem[Schmidhuber et~al.(1997)Schmidhuber, Hochreiter, et~al.]{schmidhuber1997long}
J{\"u}rgen Schmidhuber, Sepp Hochreiter, et~al.
\newblock Long short-term memory.
\newblock \emph{Neural Computation}, 9\penalty0 (8):\penalty0 1735--1780, 1997.

\bibitem[Gers et~al.(2000)Gers, Schmidhuber, and Cummins]{gers2000learning}
Felix~A Gers, J{\"u}rgen Schmidhuber, and Fred Cummins.
\newblock Learning to forget: Continual prediction with lstm.
\newblock \emph{Neural computation}, 12\penalty0 (10):\penalty0 2451--2471, 2000.

\bibitem[Galkin(2024)]{quickselect}
Michael Galkin.
\newblock {Graph \& Geometric ML in 2024: Where We Are and What’s Next (Part II — Applications)}, 2024.
\newblock URL \url{https://towardsdatascience.com/graph-geometric-ml-in-2024-where-we-are-and-whats-next-part-ii-applications-1ed786f7bf63}.

\bibitem[Veli{\v{c}}kovi{\'c} et~al.(2020)Veli{\v{c}}kovi{\'c}, Buesing, Overlan, Pascanu, Vinyals, and Blundell]{velivckovic2020pointer}
Petar Veli{\v{c}}kovi{\'c}, Lars Buesing, Matthew Overlan, Razvan Pascanu, Oriol Vinyals, and Charles Blundell.
\newblock Pointer graph networks.
\newblock \emph{Advances in Neural Information Processing Systems}, 33:\penalty0 2232--2244, 2020.

\bibitem[Xu et~al.(2018)Xu, Hu, Leskovec, and Jegelka]{xu2018powerful}
Keyulu Xu, Weihua Hu, Jure Leskovec, and Stefanie Jegelka.
\newblock How powerful are graph neural networks?
\newblock \emph{arXiv preprint arXiv:1810.00826}, 2018.

\bibitem[Zaheer et~al.(2017)Zaheer, Kottur, Ravanbakhsh, Poczos, Salakhutdinov, and Smola]{zaheer2017deep}
Manzil Zaheer, Satwik Kottur, Siamak Ravanbakhsh, Barnabas Poczos, Russ~R Salakhutdinov, and Alexander~J Smola.
\newblock Deep sets.
\newblock \emph{Advances in neural information processing systems}, 30, 2017.

\bibitem[Wagstaff et~al.(2022)Wagstaff, Fuchs, Engelcke, Osborne, and Posner]{wagstaff2022universal}
Edward Wagstaff, Fabian~B Fuchs, Martin Engelcke, Michael~A Osborne, and Ingmar Posner.
\newblock Universal approximation of functions on sets.
\newblock \emph{Journal of Machine Learning Research}, 23\penalty0 (151):\penalty0 1--56, 2022.

\bibitem[Engelmayer et~al.(2024)Engelmayer, Georgiev, and Veli{\v{c}}kovi{\'c}]{engelmayer2024parallel}
Valerie Engelmayer, Dobrik~Georgiev Georgiev, and Petar Veli{\v{c}}kovi{\'c}.
\newblock Parallel algorithms align with neural execution.
\newblock In \emph{Learning on Graphs Conference}, pages 31--1. PMLR, 2024.

\bibitem[Wang et~al.(2024)Wang, Elhag, Jaitly, Susskind, and Bautista]{wangswallowing}
Yuyang Wang, Ahmed~AA Elhag, Navdeep Jaitly, Joshua~M Susskind, and Miguel~{\'A}ngel Bautista.
\newblock Swallowing the bitter pill: Simplified scalable conformer generation.
\newblock In \emph{Forty-first International Conference on Machine Learning}, 2024.

\bibitem[Abramson et~al.(2024)Abramson, Adler, Dunger, Evans, Green, Pritzel, Ronneberger, Willmore, Ballard, Bambrick, et~al.]{abramson2024accurate}
Josh Abramson, Jonas Adler, Jack Dunger, Richard Evans, Tim Green, Alexander Pritzel, Olaf Ronneberger, Lindsay Willmore, Andrew~J Ballard, Joshua Bambrick, et~al.
\newblock Accurate structure prediction of biomolecular interactions with alphafold 3.
\newblock \emph{Nature}, pages 1--3, 2024.

\bibitem[Ong and Veli{\v{c}}kovi{\'c}(2022)]{ong2022learnable}
Euan Ong and Petar Veli{\v{c}}kovi{\'c}.
\newblock Learnable commutative monoids for graph neural networks.
\newblock In \emph{Learning on Graphs Conference}, pages 43--1. PMLR, 2022.

\bibitem[Hamilton et~al.(2017)Hamilton, Ying, and Leskovec]{hamilton2017inductive}
Will Hamilton, Zhitao Ying, and Jure Leskovec.
\newblock Inductive representation learning on large graphs.
\newblock \emph{Advances in neural information processing systems}, 30, 2017.

\bibitem[Murphy et~al.(2018)Murphy, Srinivasan, Rao, and Ribeiro]{murphy2018janossy}
Ryan~L Murphy, Balasubramaniam Srinivasan, Vinayak Rao, and Bruno Ribeiro.
\newblock Janossy pooling: Learning deep permutation-invariant functions for variable-size inputs.
\newblock \emph{arXiv preprint arXiv:1811.01900}, 2018.

\bibitem[Roy et~al.(2021)Roy, De, and Chakrabarti]{roy2021adversarial}
Indradyumna Roy, Abir De, and Soumen Chakrabarti.
\newblock Adversarial permutation guided node representations for link prediction.
\newblock In \emph{Proceedings of the AAAI conference on artificial intelligence}, volume~35, pages 9445--9453, 2021.

\bibitem[Cho et~al.(2014)Cho, van Merrienboer, Gulcehre, Bougares, Schwenk, and Bengio]{cho2014learning}
Kyunghyun Cho, B~van Merrienboer, Caglar Gulcehre, F~Bougares, H~Schwenk, and Yoshua Bengio.
\newblock Learning phrase representations using rnn encoder-decoder for statistical machine translation.
\newblock In \emph{Conference on Empirical Methods in Natural Language Processing (EMNLP 2014)}, 2014.

\bibitem[Bevilacqua et~al.(2023)Bevilacqua, Nikiforou, Ibarz, Bica, Paganini, Blundell, Mitrovic, and Veli{\v{c}}kovi{\'c}]{bevilacqua2023neural}
Beatrice Bevilacqua, Kyriacos Nikiforou, Borja Ibarz, Ioana Bica, Michela Paganini, Charles Blundell, Jovana Mitrovic, and Petar Veli{\v{c}}kovi{\'c}.
\newblock Neural algorithmic reasoning with causal regularisation.
\newblock In \emph{International Conference on Machine Learning}, pages 2272--2288. PMLR, 2023.

\bibitem[Diao and Loynd(2022)]{diao2022relational}
Cameron Diao and Ricky Loynd.
\newblock Relational attention: Generalizing transformers for graph-structured tasks.
\newblock \emph{arXiv preprint arXiv:2210.05062}, 2022.

\bibitem[Bohde et~al.(2024)Bohde, Liu, Saxton, and Ji]{bohde2024markov}
Montgomery Bohde, Meng Liu, Alexandra Saxton, and Shuiwang Ji.
\newblock On the markov property of neural algorithmic reasoning: Analyses and methods.
\newblock \emph{arXiv preprint arXiv:2403.04929}, 2024.

\bibitem[Mahdavi et~al.(2022)Mahdavi, Swersky, Kipf, Hashemi, Thrampoulidis, and Liao]{mahdavi2022towards}
Sadegh Mahdavi, Kevin Swersky, Thomas Kipf, Milad Hashemi, Christos Thrampoulidis, and Renjie Liao.
\newblock Towards better out-of-distribution generalization of neural algorithmic reasoning tasks.
\newblock \emph{arXiv preprint arXiv:2211.00692}, 2022.

\bibitem[Rodionov and Prokhorenkova(2024)]{rodionov2024discrete}
Gleb Rodionov and Liudmila Prokhorenkova.
\newblock Discrete neural algorithmic reasoning.
\newblock \emph{arXiv preprint arXiv:2402.11628}, 2024.

\bibitem[Gilmer et~al.(2017)Gilmer, Schoenholz, Riley, Vinyals, and Dahl]{gilmer2017neural}
Justin Gilmer, Samuel~S Schoenholz, Patrick~F Riley, Oriol Vinyals, and George~E Dahl.
\newblock Neural message passing for quantum chemistry.
\newblock In \emph{International conference on machine learning}, pages 1263--1272. PMLR, 2017.

\bibitem[Csord{\'a}s et~al.(2021)Csord{\'a}s, Irie, and Schmidhuber]{csordas2021neural}
R{\'o}bert Csord{\'a}s, Kazuki Irie, and J{\"u}rgen Schmidhuber.
\newblock The neural data router: Adaptive control flow in transformers improves systematic generalization.
\newblock \emph{arXiv preprint arXiv:2110.07732}, 2021.

\bibitem[Haviv et~al.(2022)Haviv, Ram, Press, Izsak, and Levy]{haviv2022transformer}
Adi Haviv, Ori Ram, Ofir Press, Peter Izsak, and Omer Levy.
\newblock Transformer language models without positional encodings still learn positional information.
\newblock \emph{arXiv preprint arXiv:2203.16634}, 2022.

\end{thebibliography}

\appendix

\section{RNAR ablations on base model and positional information}
\label{app:ablation}

In this Appendix, we aim to answer two key questions about the choice of RNAR base model, as well as its reliance on positional information---both enumerated in Table \ref{tab:compare-mp-nope}. The results are taken across only the algorithms where RNAR does not run out of memory, in order to ensure a meaningful ablation of capabilities.

\begin{table}[ht]
    \centering
    
   \caption{Test results of RNAR, RNAR-MPNN (an MPNN-based architecture with a recurrent aggregator), and RNAR-NoPE (RNAR without positional inputs) on all algorithms where RNAR does not run out of memory.}
    \begin{tabular}{lccc}
    \toprule
    \textbf{Algorithm} & \textbf{RNAR} & \textbf{RNAR-MPNN} & \textbf{RNAR-NoPE}\\\midrule
Activity Selector & $ 95.23\% \pm  0.71$ & $ 87.35\% \pm  4.19$ & $\textbf{ 96.86}\% \pm  0.31$\\
Bellman-Ford & $ \textbf{96.00}\% \pm  0.38$ & $ 93.54\% \pm  0.92$ & $ 95.33\% \pm  0.10$\\
BFS & $\textbf{ 100.00}\% \pm  0.00$ & $\textbf{ 100.00}\% \pm  0.00$ & $ 89.32\% \pm  3.96$\\
Binary Search & $\textbf{ 64.71}\% \pm  6.79$ & $ 52.75\% \pm  4.17$ & $ 62.40\% \pm  9.14$\\
Bubble Sort & $ 95.78\% \pm  0.40$ & $ 62.79\% \pm  9.68$ & $\textbf{ 96.86}\% \pm  0.35$\\
DAG Shortest Paths & $\textbf{ 96.40}\% \pm  1.47$ & $ 49.06\% \pm  4.98$ & $ 83.07\% \pm  6.41$\\
DFS & $ \textbf{100.0}\% \pm 0.00$ & $ 6.78\% \pm  2.30$ & $ 97.37\% \pm  2.15$\\
Dijkstra & $ 95.04\% \pm  1.62$ & $\textbf{ 96.75}\% \pm  0.49$ & $ 86.38\% \pm  6.67$\\
Find Max. Subarray  & $\textbf{ 83.53}\% \pm  2.17$ & $ 73.47\% \pm  0.79$ & $ 75.43\% \pm  5.67$\\
Floyd-Warshall & $ 27.49\% \pm  6.95$ & $ \textbf{41.33}\% \pm  4.56$ & $ 9.92\% \pm  4.50$\\
Graham Scan & $ 76.20\% \pm  4.51$ & $ 50.92\% \pm  3.04$ & $ \textbf{79.18}\% \pm  2.25$\\
Heapsort & $ 93.07\% \pm  1.03$ & $ 67.61\% \pm  10.71$ & $\textbf{ 95.41}\% \pm  0.33$\\
Insertion Sort & $ 93.00\% \pm  1.77$ & $ 87.88\% \pm  0.75$ & $\textbf{ 99.16}\% \pm  0.27$\\
Knuth-Morris-Pratt & $ 4.54\% \pm  2.60$ & $\textbf{ 23.01}\% \pm  10.60$ & $ 3.96\% \pm  1.62$\\
LCS Length & $ 66.91\% \pm  2.53$ & $ \textbf{77.46}\% \pm  2.83$ & $ 73.29\% \pm  4.12$\\
Matrix Chain Order & $ 25.12\% \pm  1.86$ & $ \textbf{43.55}\% \pm  9.14$ & $ 24.18\% \pm  1.92$\\
Minimum & $ 96.92\% \pm  0.09$ & $\textbf{ 98.08}\% \pm  1.16$ & $ 97.73\% \pm  0.38$\\
MST-Prim & $ 86.60\% \pm  4.42$ & $ 87.59\% \pm  3.52$ & $ \textbf{89.82}\% \pm  1.97$\\
Na\"{i}ve String Matcher & $\textbf{ 98.95}\% \pm  0.42$ & $ 28.55\% \pm  21.13$ & $ 18.81\% \pm  9.13$\\
Optimal BST & $ 36.04\% \pm  12.55$ & $ \textbf{42.29}\% \pm  13.39$ & $ 22.31\% \pm  13.06$\\
Quickselect & $ \textbf{87.08}\% \pm  2.21$ & $83.90\% \pm  3.11$ & $ 79.67\% \pm  5.54$\\
Quicksort & $\textbf{ 94.73}\% \pm  0.63$ & $ 71.57\% \pm  3.34$ & $ 94.66\% \pm  0.43$\\
Segments Intersect & $ 97.30\% \pm  0.29$ & $\textbf{ 97.84}\% \pm  0.15$ & $ 97.03\% \pm  0.21$\\
SCC & $ \textbf{48.43}\% \pm  8.01$ & $ 28.53\% \pm  3.01$ & $ 45.35\% \pm  11.01$\\
Task Scheduling & $ \textbf{88.08}\% \pm  1.30$ & $ 81.65\% \pm  0.59$ & $ 87.89\% \pm  1.34$\\
Topological Sort & $ 74.00\% \pm  8.18$ & $ \textbf{81.98}\% \pm  14.07$ & $ 76.29\% \pm  9.01$\\
\midrule
Overall average  & $\textbf{ 77.74}\%$ & $ 66.01\%$ & $ 72.22\%$\\ \bottomrule
\end{tabular}
    \label{tab:compare-mp-nope}
    \end{table}

The ablations are as follows:

\paragraph{RNAR gains additional power through architectural choices}

As already previously discussed, RNAR is not the first GNN architecture to use a recurrent aggregator---GraphSAGE \citep{hamilton2017inductive} being a notable early example. However, we remark that RNAR is not the same architecture as GraphSAGE---its base model relies on architectural novelties such as triplet messages \citep{dudzik2022graph} and gating mechanisms \citep{csordas2021neural}, as described by \citet{ibarz2022generalist}.

We argue that these improvements further amplify the impact of a recurrent aggregator, and compare against RNAR-MPNN, a model with a recurrent aggregator which otherwise uses a more typical message-passing neural network \citep{gilmer2017neural,velivckovic2022clrs} as the base model. Such an architecture is similar to GraphSAGE-LSTM.

As is evident in Table \ref{tab:compare-mp-nope}, introducing the architectural changes results in an $11\%$ increase in average OOD execution performance, which is evidence in support of our hypothesis. That being said, the simpler MPNN architecture does appear to better align with certain classes of algorithms, such as dynamic programming (see LCS Length, Matrix Chain Order and Optimal BST) as well as the Floyd-Warshall and Knuth-Morris-Pratt algorithms; these discrepancies may warrant further investigation into the optimal base model for RNAR.

\paragraph{RNAR can meaningfully exploit positional features}

Since nodes are fed into the LSTM aggregator of RNAR in a canonical order (using its positional feature), it may be argued that the model does not need to use this feature anymore. We believe that positional information may still be quite relevant, especially in graph algorithms where nontrivial tiebreaking is common. To assess this, we compare RNAR against a variant which does not use the positional feature (akin to NoPE \citep{haviv2022transformer}).

Once again, the results in Table \ref{tab:compare-mp-nope} provide evidence for our claim, with RNAR losing $5\%$ average performance when the positional features are withheld. That being said, this removal does seem to provide meaningful uplift on nearly all \emph{sorting algorithms}, which may provide useful motivation for future investigation into how the positional feature may be (mis)used by models like RNAR.

\section{RNAR's robustness against choice of node permutation}
\label{app:janossy}

One of the motivating factors for the suitability of recurrent aggregators in NAR is the fact that nodes may often have a \emph{canonical ordering} when executing algorithms---and this is certainly the case in CLRS-30. We now seek to investigate how relevant is this canonicalisation to the model's performance, by checking how well it performs when node permutations are consistently \emph{randomly} sampled, during both training and inference.

It might be noted that aggregating across randomly sampled permutations is exactly one of the strategies employed by Janossy pooling \citep{murphy2018janossy} to achieive permutation equivariance in expectation. This motivates our comparison in Table \ref{tab:compare_janossy}, where we benchmark RNAR (using canonical node order) against RNAR-Janossy-$k$, which chooses $k$ random permutations, runs the LSTM over each of them, and averages the resulting embedding vectors. We focus on values of $k\in\{1, 2, 3\}$, to provide a meaningful indication of trends without requiring too many computational resources. 

As in Appendix \ref{app:ablation}, we only take results across the algorithms where RNAR-Janossy-3 does not run out of memory, in order to ensure a meaningful ablation.

\setlength{\tabcolsep}{2pt}
\begin{table}[ht]
    \centering
   \caption{Test results of RNAR and RNAR-Janossy-$k$ (where embeddings are aggregated across $k$ random node permutations) models on all algorithms where these ablations do not run out of memory.}
    \begin{tabular}{lcccc}
    \toprule
    \textbf{Algorithm} & \textbf{RNAR} & \textbf{RNAR-Janossy-1} & \textbf{RNAR-Janossy-2} & \textbf{RNAR-Janossy-3} \\\midrule
Activity Selector & $ 95.23\% \pm  0.71$ & $ 95.54\% \pm  0.93$ & $ 95.65\% \pm  0.64$ & $ \textbf{96.13}\% \pm  0.99$\\
Bellman-Ford & $ 96.00\% \pm  0.38$ & $ 96.31\% \pm  0.43$ & $\textbf{ 96.64}\% \pm  0.05$ & $ 96.22\% \pm  0.40$\\
BFS & $\textbf{ 100.00}\% \pm  0.00$ & $\textbf{ 100.00}\% \pm  0.00$ & $\textbf{ 100.00}\% \pm  0.00$ & $\textbf{ 100.00}\% \pm  0.00$\\
Binary Search & $\textbf{ 64.71}\% \pm  6.79$ & $ 55.40\% \pm  6.96$ & $ 36.37\% \pm  4.02$ & $ 60.60\% \pm  7.16$\\
DAG Shortest Paths & $\textbf{ 96.40}\% \pm  1.47$ & $ 89.71\% \pm  4.17$ & $ 80.90\% \pm  6.16$ & $ 84.40\% \pm  5.76$\\
DFS & $ \textbf{100.0}\% \pm  0.00$ & $ 17.52\% \pm  4.11$ & $ 18.98\% \pm  2.25$ & $ 39.53\% \pm  9.44$\\
Dijkstra & $ 95.04\% \pm  1.62$ & $ 89.67\% \pm  5.44$ & $ 92.33\% \pm  1.24$ & $ \textbf{96.54}\% \pm  1.09$\\
Find Max. Subarray  & $\textbf{ 83.53}\% \pm  2.17$ & $ 67.93\% \pm  7.39$ & $ 78.64\% \pm  1.02$ & $ 79.14\% \pm  2.28$\\
Floyd-Warshall & $ 27.49\% \pm  6.95$ & $\textbf{ 70.36}\% \pm  10.24$ & $ 46.70\% \pm  18.75$ & $ 47.86\% \pm  17.72$\\
Graham Scan & $ 76.20\% \pm  4.51$ & $ 83.59\% \pm  4.75$ & $ 86.61\% \pm  3.22$ & $\textbf{ 90.88}\% \pm  0.78$\\
Insertion Sort & $ 93.00\% \pm  1.77$ & $ 94.72\% \pm  0.56$ & $ \textbf{97.44}\% \pm  0.86$ & $ 95.01\% \pm  1.39$\\
Knuth-Morris-Pratt & $ 4.54\% \pm  2.60$ & $ 0.67\% \pm  0.28$ & $ \textbf{8.98}\% \pm  2.54$ & $ 3.65\% \pm  1.08$\\
LCS Length & $ 66.91\% \pm  2.53$ & $ 82.97\% \pm  2.00$ & $\textbf{ 84.78}\% \pm  0.07$ & $ 78.84\% \pm  2.98$\\
Matrix Chain Order & $ 25.12\% \pm  1.86$ & $ 84.94\% \pm  2.79$ & $\textbf{ 86.59}\% \pm  3.35$ & $ 82.97\% \pm  4.30$\\
Minimum & $ 96.92\% \pm  0.09$ & $ \textbf{97.20}\% \pm  0.29$ & $ 93.81\% \pm  1.76$ & $ 85.63\% \pm  9.86$\\
MST-Prim & $ 86.60\% \pm  4.42$ & $ 91.59\% \pm  0.96$ & $\textbf{ 91.72}\% \pm  1.12$ & $ 90.93\% \pm  2.81$\\
Na\"{i}ve String Matcher & $\textbf{ 98.95}\% \pm  0.42$ & $ 5.63\% \pm  2.30$ & $ 61.40\% \pm  24.26$ & $ 11.75\% \pm  5.74$\\
Optimal BST & $ 36.04\% \pm  12.55$ & $ 50.21\% \pm  20.45$ & $\textbf{ 78.06}\% \pm  3.40$ & $ 70.51\% \pm  9.91$\\
Segments Intersect & $ \textbf{97.30}\% \pm  0.29$ & $ 92.17\% \pm  1.57$ & $ 91.45\% \pm  2.14$ & $ 94.22\% \pm  1.60$\\
Task Scheduling & $ 88.08\% \pm  1.30$ & $ 88.18\% \pm  0.53$ & $\textbf{ 88.55}\% \pm  0.91$ & $ 87.74\% \pm  0.96$\\
Topological Sort & $ 74.00\% \pm  8.18$ & $\textbf{ 95.13}\% \pm  1.11$ & $ 73.08\% \pm  12.39$ & $ 74.55\% \pm  8.79$\\
\midrule
Overall average  & $ \textbf{76.29}\%$ & $ 73.78\%$ & $75.65\%$ & $ 74.63\%$\\\bottomrule
\end{tabular}
    \label{tab:compare_janossy}
    \end{table}

While our results in Table \ref{tab:compare_janossy} do indicate a slight average advantage to the canonical order used by RNAR, the comparison is substantially less clear-cut; there are several algorithms from which there is a very clear uplift from regularising RNAR towards permutation equivariance in this way. If we take into consideration the number of algorithms where Janossy pooling runs out of memory, it may still be concluded that canonicalisation is the better choice. That being said, our ablation points at a clear line of future work, which would study principled ways to regularise NARs without symmetries (such as permutation equivariance) towards satisfying these symmetries in expectation.

\section{Timing performance of RNAR}
\label{app:timing}

\begin{figure}
    \centering
    \includegraphics[width=\linewidth]{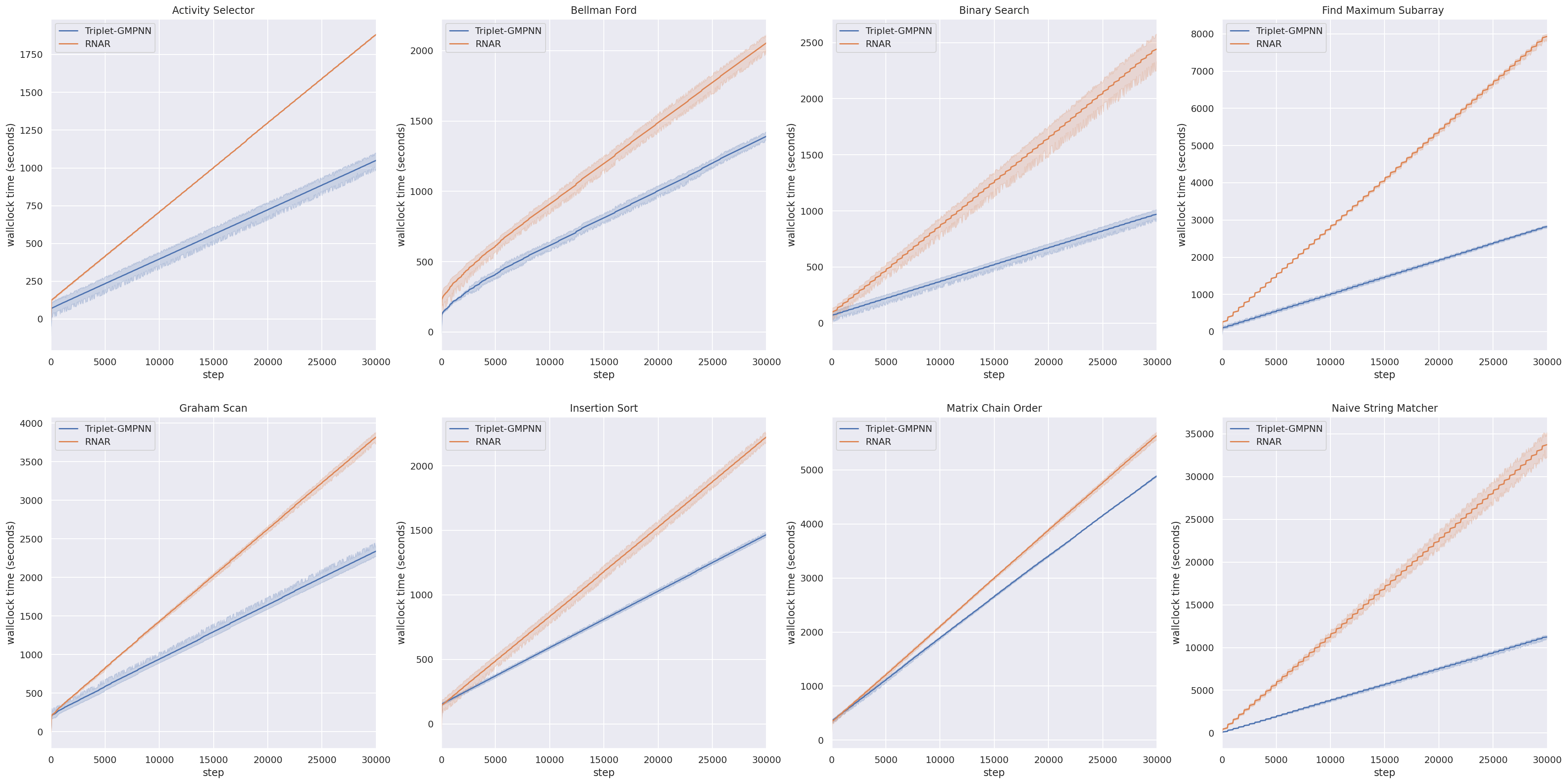}
    \caption{A comparison of the wall-clock times ($y$-axis, in seconds) required for completing a certain number of training steps ($x$-axis), for the baseline Triplet-GMPNN (in blue) against RNAR (in orange), across eight representative algorithms in CLRS-30.}
    \label{fig:timing}
\end{figure}

In Figure \ref{fig:timing} we show the effects of adding RNAR on computation time compared to the baseline Triplet-GMPNN. As expected, the overall training steps-per-second is worse affected for algorithms that require more intermediate iterations before arriving at the final answer.

\end{document}